\documentclass[11pt,letterpaper]{article}

\usepackage[margin=1in]{geometry}
\usepackage[T1]{fontenc}
\usepackage{lmodern}
\usepackage{microtype}
\usepackage{amsmath,amssymb,amsthm}
\usepackage{bbm}
\usepackage{bm}
\usepackage{graphicx}
\usepackage[x11names]{xcolor}
\usepackage{xspace}
\usepackage[ruled]{algorithm2e}
\usepackage[normalem]{ulem}
\usepackage[authoryear,round]{natbib}
\usepackage{url}
\usepackage{hyperref}
\hypersetup{
  colorlinks=true,
  linkcolor=blue,
  citecolor=blue,
  urlcolor=magenta,
  pdftitle={Towards a Theory of Learnability via Debate},
  pdfauthor={Tosca Lechner, Vinayak Pathak, Sivan Sabato, Ruth Urner}
}
\bibpunct{(}{)}{;}{a}{,}{,}

\newtheorem{theorem}{Theorem}
\newtheorem{lemma}[theorem]{Lemma}
\theoremstyle{definition}
\newtheorem{definition}[theorem]{Definition}
\newtheorem{example}{Example}
\theoremstyle{plain}
\newtheorem{claim}[theorem]{Claim}

  \newcommand{\Ccal}{{\mathcal C}}
  \newcommand{\C}{{\Ccal}}

  \newcommand{\Hcal}{{\mathcal H}}
  \renewcommand{\H}{{\Hcal}}

  \newcommand{\V}{{\mathcal V}}

  \newcommand{\X}{{\mathcal X}}
  
  \newcommand{\Y}{{\mathcal Y}}

  \newcommand{\tLo}{{{\mathcal L}_P}}
  \newcommand{\fLo}{{{\mathcal L}_R}}
  
  \newcommand{\tRe}{{{\mathcal R}_P}}
  \newcommand{\fRe}{{{\mathcal R}_R}}
  \newcommand{\Mi}{{\mathcal M}}
  \newcommand{\mi}{\mathrm{m}}

 \newcommand{\ex}{\mathbb{E}}
 
 \newcommand{\naturals}{\mathbb{N}}

\newcommand{\indct}{\mathbbm{1}}

  \newcommand{\dt}{\texttt{Proposer}\xspace}
  \newcommand{\df}{\texttt{Responder}\xspace}

  \renewcommand{\c}{c}
  \newcommand{\ct}{{p}}
  \newcommand{\cf}{{r}}
  \renewcommand{\sc}{\bar{c}}
  \newcommand{\hy}{{\hat{y}}}
  \newcommand{\vy}{{\widetilde{y}}}
  \newcommand{\ty}{y}

  \newcommand{\T}{{\mathcal T}}
  \newcommand{\depth}{\mathrm{depth}}
  \newcommand{\dldim}{\mathrm{DLdim}}
 \newcommand{\debatedimname}{debate learning dimension\xspace}

\DeclareMathOperator*{\argmin}{arg\,min}
\DeclareMathOperator*{\argmax}{arg\,max}

\title{Towards a Theory of Learnability via Debate}
\author{
  Tosca Lechner\textsuperscript{1}\quad
  Vinayak Pathak\textsuperscript{2,}\thanks{corresponding author: \href{mailto:path.vinayak@gmail.com}{\nolinkurl{path.vinayak@gmail.com}}}\quad
  Sivan Sabato\textsuperscript{1,3,\textdagger}\quad
  Ruth Urner\textsuperscript{1,4}\\[0.6em]
  {\small\textsuperscript{1}Vector Institute}\\
  {\small\textsuperscript{2}Computational Rational Agents Laboratory (CORAL)}\\
  {\small\textsuperscript{3}McMaster University, Department of Computing and Software}\\
  {\small\textsuperscript{4}York University, EECS Department}\\[0.2em]
  {\footnotesize\textsuperscript{\textdagger}Canada CIFAR AI Chair}
}
\date{}
\begin{document}

\maketitle

\begin{abstract}%
  Debate has been recently proposed as a training protocol that is hoped to
produce more aligned models even as model capabilities scale. The idea is
that the human, instead of providing supervision in the form of labels,
judges a debate between two models that compete to prove the correctness of
opposite labels. It can be shown that functions of a much higher computational complexity  can be
correctly judged in this way, compared to the complexity of functions that can be directly evaluated using a computationally weak evaluator. This suggests that perhaps debate can be
used to teach very complicated functions to models. However, previous work
has largely ignored the fact that this is a \emph{training} protocol, and
simply analyzes the game being played between the two debaters for a known
judge. We take the first steps towards understanding the learning problem
implicit within the protocol, by formulating it as an online learning
problem. We show that once we formulate the learning problem precisely,
there are cases where the best way to learn given access to a judge is
different from running an adversarial debate. This motivates us to define a
cooperative version of the protocol, whose learnability we characterize
using a combinatorial dimension. We also consider the possibility of the
human judge being fallible and occasionally judging the debate incorrectly.
We show that a small number of errors in judgement can lead to a large error
in learning.
\end{abstract}


\section{Introduction}
As AI models grow in capabilities, how do we ensure that they continue to act in the interest of humans? A long line of work~\citep{yudkowsky2008artificial,soares2015corrigibility,russell2015research,amodei2016concrete,everitt2017corrupted,soares2017agent,armstrong2018occam,soares2018value,hubinger2019risks,taylor2020alignment,skalse2022reward} has argued that this is a hard problem, and recent events such as the infamous HuggingFace incident~\citep{openai2026huggingface} have demonstrated the difficulties in practice. 

One way to study this problem more formally has been termed \emph{scalable oversight}. Here, it is assumed that our true preferences are encoded by a function $f: \X \rightarrow \Y$ that is hard for us to compute, but easy for an AI. For example, suppose $x \in \X$ encodes the current state of the world and $y \in \Y$ is the desired action. If this involves long-term planning then $f$ can be hard to compute for us. There are many reasons for why an AI can be more powerful than us. The scalable oversight literature assumes that the reason is largely computational. For example, early works assumed that $f$ was a PSPACE-complete problem and that an AI was able to solve it~\citep{irving2018debate}, while a human was only able to solve problems in P. Some more recent works assume other kinds of computational superiority~\citep{browncohen2024debate,browncohen2025prover}. Since we are not able to compute $f$ ourselves, it is unclear how we can: a) design a training process that teaches a model to compute $f$, and b) check that a trained AI is indeed computing $f$. Scalable oversight aims to provide an answer. However, the currently existing literature has mostly focused on (b) while largely ignoring (a). 

That (b) might be achievable can be seen using the following analogy that appeared in~\cite{irving2018debate}. Suppose $x$ represents a SAT instance and $y$ denotes whether it is satisfiable. To prove to the human that $y=\mathrm{true}$, the AI just needs to provide the satisfying assignment along with $y$. In general, the scalable oversight literature assumes that $f$ sits within some level of the polynomial hierarchy, thus providing a means to certify whether AI's solution is correct. For example, suppose $f$ lies in the intersection of the two classes at level $2n$ of the polynomial hierarchy (at level 1, these are NP and co-NP).\footnote{Formally, this is the class $\Sigma_{2n}^{\mathrm{P}} \cap \Pi_{2n}^{\mathrm{P}}$.} Then there exists a polynomial time verifier $v$ such that for all $x$, the correct answer is $y$ iff $\exists c_1 \forall c_2 \ldots \exists c_{2n-1} \forall c_{2n}\, v(x,y,c_1,\ldots ,c_{2n}) = \mathrm{true}$. This motivates the \emph{debate} protocol~\citep{irving2018debate,browncohen2024debate}, where the AI is thought of as a pair of \emph{debaters} (also implemented as AI models). Given $x$, the \dt proposes a $y$ and tries to defend it. The \df argues for the opposite label. The two debaters  generate alternating arguments $c_1,c_2,\ldots , c_{2n-1},c_{2n}$, and the human judge $v$ judges whether the debate transcript proves that the answer to $x$ is $y$. The \dt's goal is to make the verdict `yes' and the \df's goal is the opposite. Due to the definition of this complexity class, one has the guarantee that the \dt has a winning strategy for a given $x$ if and only if its proposed label is equal to $f(x)$. If we allow the number of rounds of the debate to grow polynomially with the size of $x$, the problems whose answers can be certified this way are exactly those in PSPACE, and thus it is tempting to say that a human judge could potentially supervise models that solve problems in PSPACE (which was indeed the motivation presented in the first paper on debate in \citealp{irving2018debate}).

However, as mentioned above, the current literature ignores the important part (a), namely, how do we ensure that a model \emph{learns} the correct function $f$. Just because the existence of \dt's winning strategy is equivalent to the \dt returning the correct label, it does not imply that the winning strategy can be found. Moreover, it is likely that the advantage that an AI might have over humans is not as much about computation as about the ability to learn facts about the universe that humans haven't learned. Indeed, according to the extended Church-Turing thesis~\citep{vergis1986complexity, aaronson2005physical}, every phyically realizable computer can be \emph{efficiently} simulated by a (quantum) Turing machine. Thus, a computational intelligence, no matter how ``superhuman'' it gets, is unlikely to be solving worst case instances of NP-hard problems, let alone PSPACE-hard problems. If an AI is seemingly able to solve NP-hard problems (such as AlphaFold~\citep{jumper2021alphafold}), it is likely because it has \emph{learned} something about the distribution from which the universe is creating the input instances. Thus, we claim that learning should be at the centre of any investigation into scalable oversight.

\subsection{Our results}
We model debate as an online learning problem. Nature produces a sequence of $x_t$'s, and at each round $t$, two debaters produce a sequence $\sc_t$ of certificates. The human is modeled by a verifier $v^*$ as above. Thus, given $x_t$, a proposed $\hy_t$, and a debate transcript $\sc_t$, $v^*$ can decide whether the transcript proves that $\hy_t$ is the correct label for $x_t$ and provides this feedback to the learner. Each verifier $v$ induces a ground truth hypothesis $h$. The debaters need to learn the true $h^*$. They don't know the true $v^*$ (because otherwise they would also know $h^*$) but know a class $\V$ that $v^*$ lies in. A \emph{mistake} is when the label $\hy_t$ proposed by the \dt is different from the true label $h^*(x_t)$. In the \emph{realizable} case, we assume that  the feedback is consistent with the true $v^*$ at each round.

Modeled this way, the learning problem becomes explicit and already opens up intriguing questions: First, what objective should adversarial debaters optimize? We show that the loss objective from previous work is not as useful in a learning setting, and propose a regret objective instead. Second, is it always better for the debaters to play adversarially? For example, can there be a case where the learning problem is solved more quickly if the two debaters cooperate? We answer this in the affirmative, where we show that there exist instances of the debate problem where learning is impossible when the debaters act adversarially to each other, but can be done with a constant number of mistakes if they cooperate. These results are presented in Section~\ref{sec:oppositional}.

Motivated by this example, we define the \emph{cooperative} version of the problem,
in which the debaters work towards the joint goal of using access to $v$ in the most efficient way in order to learn the ground truth. We present a characterization of the optimal mistake bound by (deterministic) cooperative debaters in the realizable setting (Section~\ref{sec:cooperative-case}), using a new combinatorial dimension that we define, termed \emph{\debatedimname}. A significant challenge in this setting is that in contrast with standard online learning protocols, the learner never observes the true labels that it is trying to predict. 

A common criticism of the debate protocol is that it assumes the infallibility of humans~\citep{irving2019social,browncohen2024debate,parrish2022single,parrish2022two}. It is unclear that a human can implement such a $v$ that aligns with the ground truth on all $x$. Thus, we also consider an unrealizable version of the problem, in which we assume that $v$'s output can be arbitrarily corrupted during some of the rounds. We call these rounds \emph{misspecified}, and show that for every constant $d > 1$, there are instances of the problem such that in the realizable setting, the optimal deterministic mistake bound is $d$, while if there are $k$ misspecified rounds, 
then every learner makes at least $\Omega(dk)$ mistakes. This contrasts with Online Learning, in which any hypothesis class with a finite Littlestone dimension $L$ admits a ``no-regret'' algorithm for the $k$-non-realizable setting with a mistake upper bound of $k +O(\sqrt{k\cdot L} + L)$, implying that the prediction error rate converges to the misspecification rate \citep{FilmusHaMeMo23,BenDavidPaSh09,AlonBeDaMoNaYo21}. 
Our lower bound for debate learning shows that debate learning is significantly more  sensitive to misspecification errors: even a small rate of misspecification errors can induce an unreasonably large prediction error, and the error rate may never approach the misspecification rate. 

For some of our results, the proofs are deferred to the appendix.

\subsection{Related work}
There is very little previous work that analyzes learning in the
setting of debate. Perhaps the two that are closest to our work
are~\citet{chen2023playing} and~\citet{buzaglo2025hidden}. Both model
debate as a repeated game between the two debaters with a fixed payoff
matrix. Since the space of arguments is too large to enumerate, the
debaters access this matrix through an optimization oracle, and both
papers give efficient no-regret strategies under this access. By
classic results, the time-averaged play then converges to an
equilibrium~\citep{freund1999adaptive,hart2000simple}. In this model,
the debaters never need to learn the verifier, and it is assumed that the equilibrium answer is correct.
In our model, the verifier is unknown, each round reveals a single
verdict on a new input, and we measure success by the number of
mistakes with respect to the ground truth.

\section{Formal Setup}\label{sec:formal}
We let $\X$ denote a \emph{domain}, $\Y = \{0,1\}$ a binary \emph{label space}, and a finite $\C$ a \emph{certificate space}.  
A \emph{hypothesis} is a function $h:\X\to\Y$, and we use $\H$ to denote a \emph{hypothesis class}.
A \emph{debate of length} $n$ is a sequence $(\c_1, \c_2, \ldots, \c_{2n})$ of $2n$ certificates $c_i\in\C$. We think of these certificates as being produced alternately by two debaters. That is, we think of the certificates $c_i$ of odd index as being produced by the first debater, which we will call the \dt, and the certificates $c_i$ of even index by the second debater, which we will refer to as the \df. To clarify the role of the arguments in the debate, we usually will use the notation $(\ct^1, \cf^1, \ct^2, \cf^2, \ldots,  \ct^n, \cf^n)$ for a debate of length $n$

The debate is moderated through a verifier, which takes in a domain point and a debate of length $n$ and produces a label for $x$ given the arguments in the debate. 
The following definition specifies the relationship between hypotheses and verifiers in our framework.

\begin{definition}\label{def:verifier_validates_hypothesis}
Let $h:\X\to\Y$ be a hypothesis. We say that verifier $v: \X\times \C^{2n} \to \Y$ \emph{validates} hypothesis $h$ through debate of length $n$ if for all $x\in\X$ there exists a unique $y\in \Y$ such that
\begin{align*}
\exists\ct^1\in\C\,\forall\cf^1\in\C\,\exists\ct^2\in\C\,\forall \cf^2\in\C\ldots\,\exists\ct^n\in\C\,\forall\cf^n\in\C,
v(x, \ct^1, \cf^1, \ct^2, \cf^2, \ldots,  \ct^n, \cf^n) = y,
\end{align*}
and $h(x) = y$ for this $y$. 
In this case, we also say that $v$ is a \emph{verifier} for hypothesis $h$. We call a verifier \emph{valid} if there exists a hypothesis $h$ that is validated by $v$.
\end{definition}

Note that a given valid verifier $v$ validates exactly one hypothesis $h$ in the above sense. We will use notation $h_v$ for this hypothesis when we want to make this relationship between verifier and its hypothesis explicit. On the other hand, a given hypothesis $h$ may be validated by multiple verifiers $v$.
We call a given set of valid verifiers a \emph{verifiers class}, and denote it $\V$. Further, we let $\H_\V$ denote the hypothesis class of all functions that are validated by verifiers $v\in\V$. If clear from context, we may simply write $\H$ for $\H_\V$.
For a hypothesis class $\H$ we say that $\H$ is \emph{supported by verifier class} $\V$ if $\H = \H_\V$.

\paragraph{Learning via debate} The goal of \emph{learning via debate} is to recover an underlying \emph{true hypothesis}, which we denote by $h^*$. We assume that the two debaters interact with a verifier $v^*\in\V$ that validates the true hypothesis $h^*$ (and thus $h^*\in\H = \H_\V$).
In the \emph{realizable case} all label feedback is provided by the verifier $v^*$.
In the \emph{$k$-non-realizable} case, the label feedback in at most $k$ rounds may be different from that of $v^*$. We call these rounds \emph{misspecified}. Of course, debaters \dt and \df do not know the identity of $h^*$ nor the identity of $v^*\in\V$. They only have access to the verifier class $\V$, and thereby also to its induced hypothesis class $\H_\V$.

We first give the debate protocol for the realizable setting. 
\paragraph{The debate protocol:}
\begin{itemize}
\item For rounds $t = 1\ldots T$ repeat:
\begin{enumerate}
 \item The environment poses an $x_t\in\X$
 \item Debater \dt hypothesizes label $\hy_t$ (also called ``prediction'' or ``predicted label'')
 \item For $i = 1 \ldots n$ repeat:
 \begin{itemize}
  \item Debater \dt brings argument $\ct^i_t$ (in support of label $\hy_t$)
  \item Debater \df brings argument $\cf^i_t$ (aiming to falsify label $\hy_t$)
 \end{itemize}
Let $\sc_t=(\ct^1_t,\cf^1_t,\ldots,\ct^n_t,\cf^n_t)$.
 \item The debaters receive $\vy_t=v^*(x_t,\sc_t)$. 
 \item %
       The environment records (but does not reveal) $\ty_t = h^*(x_t)$ 
\end{enumerate}
\end{itemize}

A run of the debate protocol is evaluated through the \emph{number of mistakes} in predicting the label $\ty_t$. For a fixed interaction sequence, denote the number of mistakes by
$\Mi(T) = \sum_{t = 1}^T \indct[\hy_t \neq \ty_t].$

A standard approach for using debate as a learning strategy considers the two debaters as playing the zero-sum game of winning the argument~\citep{irving2018debate,chen2023playing,browncohen2024debate,browncohen2025prover}.
Standard formalizations \citep{irving2018debate,browncohen2024debate} define a loss for each debater.\footnote{Some works let the verifier output a score rather than a binary verdict, measuring how strongly the debate transcript supports the proposed label~\citep{chen2023playing,arnesen2024training,browncohen2025prover}; we use binary verdicts throughout.} Debater \dt aims to defend its hypothesized label $\hy_t$ and is therefore evaluated through the following loss: $\tLo(T) = \sum_{t = 1}^T \indct[\vy_t \neq \hy_t].$ 
Debater \df, on the other hand, is evaluated against its aim to falsify the hypothesis that the first debater proposed, by setting its certificates in such a way that the verifier will conclude with a label $\vy_t$ that is not the hypothesized label $\hy_t$. 
Therefore, \df is evaluated through the opposite loss: 
 $\fLo(T) = \sum_{t = 1}^T \indct[\vy_t = \hy_t] = T - \tLo(T).$
If the learning problem is ignored, so that $v^*$ is known to both debaters, then a \dt optimizing $\tLo(T)$ with respect to a \df that optimizes $\fLo(T)$ leads to \dt always proposing the true label of $h^*$ verified by $v^*$, with proposer certificates that cannot be refuted. Moreover, this leads to $\Mi(T) = \tLo(T) = 0$, motivating this optimization objective. We discuss below the setting of two debaters that optimize their objectives separately, which we term the \emph{oppositional setting}. 

We can also consider \dt and \df that can be cooperative, with the joint goal to minimize $\Mi(T)$. 
In this case, \df provides its certificates in good faith ``for the sake of the argument'' so as to lead the verifier to the correct conclusions, having taken into account relevant potential counter-arguments. In this case, the two debaters can optimize their behavior together and can be thought of as a single entity that we also sometimes refer to as \emph{the learner} in this case. We study this learning setting below.

\section{Cooperative Debaters}
\label{sec:cooperative-case}
In this section, we start by exploring learning via debate using cooperative debaters. As mentioned above, we will call them jointly \emph{the learner} in this case. To analyze which classes $\V$ of verifiers allow for learning with finite mistake bound, we will define a tree structure that represents potential runs of the debate protocol for debate length $n$.

In our tree construction, edges correspond to all possible learner actions 
and nodes correspond to worst-case actions of the environment. 
This type of construction has been previously used in \cite{oz2026discriminative} for a different interactive learning protocol, Discriminative Feature Feedback. Nonetheless, as we will see below, our definition of dimension based on the tree will be quite different from standard ones, due to the special challenges of the debate protocol.

In our case, the learner's action in each round includes the label $\hy$ that the \dt poses, as well as the debate $\sc$ of length $n$ that then ensues between the two debaters.
The environment's actions include a verifier evaluation $\vy$ for some verifier consistent with the previous rounds, as well as the  domain instance that the environment poses for the next round.
The root node only records the instance that the protocol is initiated with.

\begin{definition}[Debate Tree]
Given a debate length $n$ and a verifier class $\V$, we define a \emph{length-$n$-debate tree for class $\V$} to be a directed tree $\T$ with the following properties:
\begin{enumerate}
    \item Every node $\alpha$ in $\T$ is a pair of a domain point and a label $\alpha=(x_\alpha, \vy_\alpha)$.
    \item The root node is denoted by $\rho$. Its domain point is denoted by $x_\rho$, and its label is $\vy_\rho = \sqcup$, where $\sqcup\notin\Y$. \item The root node $\rho$ is the only node in $\T$ with associated label $\vy_\rho \notin\Y$: for all nodes $\alpha\neq \rho$, we have $\vy_\alpha\in \Y$.
    \item Edges $(\alpha, \beta)$ in $\T$ are directed away from the root node, and annotated with a label $\ell^\alpha_\beta = (\hy^\alpha_\beta, \sc^\alpha_\beta)$, where $\sc^\alpha_\beta = (\ct^1_{\alpha\beta}, \cf^1_{\alpha\beta}, \ct^2_{\alpha\beta}, \cf^2_{\alpha\beta}, \ldots,  \ct^n_{\alpha\beta}, \cf^n_{\alpha\beta})$ is a debate of length $n$.
    \item For each non-leaf node $\alpha$ and each possible edge annotation $(\hy, \sc)$, there is exactly one child node $\beta$ to $\alpha$ with $\ell^\alpha_\beta = (\hy, \sc)$
    \item All directed paths in $\T$ are finite, and consistent with a verifier $v\in\V$ (in the sense defined below).
\end{enumerate}
We let $\Delta_n(\V)$ denote the set of all length-$n$-debate trees $\T$ for verifier class $\V$.
\end{definition}

Let $(\alpha,\beta)$ be an edge in the tree $\T$ and let $v\in\V$ be a verifier. We say \emph{$v$ is consistent with edge $(\alpha, \beta)$} if
\[
v(x_\alpha, \sc^\alpha_\beta) = v(x_\alpha, \ct^1_{\alpha\beta}, \cf^1_{\alpha\beta}, \ct^2_{\alpha\beta}, \cf^2_{\alpha\beta}, \ldots,  \ct^n_{\alpha\beta}, \cf^n_{\alpha\beta}) = \vy_\beta.
\]
Let $p$ be a directed path (down the tree) in $\T$. We say that the path $p$ is consistent with verifier $v$ if all edges in $p$ are consistent with $v$.

To analyze learnability, we will define \emph{version spaces} of verifiers that agree with paths in the debate tree.
 We define the \emph{verifier version space of path $p$} to be the set
\[
 \V_p = \{ v\in\V ~\mid~ v \text{ is consistent with all edges in }p\}.
\]
Recall that each verifier $v$ induces one hypothesis $h$. Therefore, a verifier version space $\V_p$ induces a corresponding hypothesis version space $\H_p$. Recall also that more than one verifier may induce the same hypothesis.

For a descendant node $\beta$ of a node $\alpha$ in $\T$, we let $p^\alpha_\beta$ denote the directed path from $\alpha$ to $\beta$.  In case $\alpha =\rho$ is the root node, we may simply use the notation $p_\beta = p^\rho_\beta$ to denote the path from root to node $\beta$. We will let $P^\alpha$ denote the set of all root-to-leaf paths in $\T$ that start at $\alpha$.

For version spaces, we may use the shorthand notations $\V^\alpha_\beta = \V_{p^\alpha_\beta}$ and $\V_\beta = \V_{p_\beta}$. That is $\V_\beta$ is the verifier version space at a node $\beta$ in the tree: $\V_\beta$ contains exactly those verifiers that are consistent with the debate protocol as run from the beginning $\rho$ to node $\beta$ in the tree. Analogously, we use $\H^\alpha_\beta$ and $\H_\beta$ for the corresponding induced hypothesis version spaces.

For any hypothesis $h:\X\to\Y$, we let $\mi(h, p )$ denote the number of disagreements between $h$ and the hypothesized labels on the edges of path $p$. That is $\mi(h, p) = \sum_{(\alpha, \beta)\in p}\indct[h(x_\alpha)\neq \hy^\alpha_\beta]$.
Note that if a verifier $v$ is consistent with path $p$ and $h_v$ is the hypothesis corresponding to $v$, then  $\mi(h_v, p)$ corresponds to the number of mistakes the debaters incur in case candidate verifier $v$ is the true $v^*$.

For a tree $\T$ and node $\alpha$ in $\T$, we let $\T^\alpha$ denote the subtree rooted at $\alpha$ in $\T$. Recall that $P^\alpha$ is the set of all root-to-leaf paths in $\T$ that are starting at $\alpha$.
We define the \emph{depth} of subtree $\T^\alpha$ with respect to verifier class $\V$ as follows:
\[
\depth(\T^\alpha, \V) = \min_{q\in P^\alpha}~ \max_{h\in \H_\alpha \cap \H_q}~  ( \mi(h, p_\alpha) + \mi(h, q) ).
\]

Note that the depth of a subtree rooted at $\alpha$ counts the number of (past and future) mistakes that the debate could incur with verifiers that are consistent with the path up to $\alpha$.
The use of past mistakes in the depth of a subtree is crucial: because the learner (the debaters) never directly finds out the true  labels $\ty_t$, it must keep track of all consistent hypotheses and its mistakes with respect to them throughout the run.
The depth of the whole tree $\T$ is simply its depth at root node $\rho$, that is $\depth(\T) = \depth(\T^\rho)$. Note that at the root node, there is no past, and thus only future mistakes are counted:
\[
\depth(\T,\V) = \depth(\T^\rho,\V) = \min_{q\in P^\rho}~ \max_{h\in \H_\V \cap \H_q}~  ( \mi(h, p_\rho) + \mi(h, q) )  = \min_{q\in P^\rho}~ \max_{h\in \H_{q}}~    \mi(h, q).
\]

With this, we are ready to define the \emph{\debatedimname} for debate length $n$ of a verifier class:
\begin{definition}[Debate Learning Dimension]%
 Let $\V$ be a verifier class for domain $\X$. Let $\Y$ be a label space, $\C$ be a certificate space, and $n\in\naturals$ be a debate length. The \emph{length-$n$-\debatedimname} $\dldim(\V, n)$ of $\V$ is the largest possible depth of a debate tree for debate length $n$:
 \[
  \dldim(\V, n) = \sup_{\T\in\Delta_n(\V)} \depth(\T,\V).
 \]
If the debate length $n$ is fixed (or clear from context) we will also use the shorthand notation $\dldim(\V) =   \dldim(\V, n)$.
\end{definition}
We will show that the \debatedimname characterizes when a hypothesis class $\H$ that is supported by a verifier class $\V$ is learnable via deterministic debate with a finite mistake bound. 

\begin{theorem}
Let $\H\subseteq \Y^\X$ be some hypothesis class. Then $\H$ is learnable from length-$n$-debate with finite mistake bound by a deterministic learner if and only if the learner has access to a verifier class $\V$ with $\H = \H_\V$ and $\dldim(\V, n) <\infty$.
Moreover, the optimal mistake bound for learning $\H$ through access to $\V$ is exactly $\dldim(\V, n)$, the length-$n$-\debatedimname of $\V$.
\end{theorem}

The theorem follows directly from the subsequent two lemmas in this section. Lemma \ref{lemma:cooperative_deterministic_lowerbound} shows that our dimension lower bounds the optimal mistake bound, while Lemma \ref{lemma:cooperative_deterministic_upperbound} provides the corresponding upper bound.
We start by showing that the \debatedimname induces a lower bound on the number of mistakes that the environment can enforce.

\begin{lemma}\label{lemma:cooperative_deterministic_lowerbound}
Let $\H\subseteq \Y^\X$ be some hypothesis class, for which there exists a verifier class $\V$ with $\H = \H_\V$ and $\dldim(\V, n) = d \in \naturals$. Then, for any deterministic choices of the debaters in the debate protocol with verifier class $\V$, the environment can enforce $d$ mistakes within finitely many rounds.
\end{lemma}

\begin{proof}
Assume $\dldim(\V, n) = d$. 
Since $d\in\naturals$ is finite, there exists a debate tree $\T$ with $\depth(\T,\V) = d$. 
From the definition of depth
\[
\depth(\T,\V) =  \min_{p\in P^\rho}~  \max_{h\in \H_p}~ \mi(h, p)
\]
it follows that for any $p \in P^\rho$, there is some $h$ that induces $d$ mistakes. Let $t_p$ denote the round after which this $h$ has induced $d$ mistakes in path $p$. Note that every $p$ in $\T$ is finite, and the fan-out of the tree $\T$ is finite (since $\C$ is finite). Hence, 
$\max_{p\in P^\rho} t_p$ is some $b \in \naturals$. 
It follows that for any sequence of outputs chosen by the learner, the environment has verifier responses and a consistent choice of $h$ that enforce $d$ mistakes in at most $b$ steps. 
\end{proof}

For the upper bound, we define the Standard Optimal Algorithm (SOA) for the length-$n$-debate protocol. Assume that the debate protocol is now in round $t$, and let $p_t$ be the path (of a debate tree) recording the previous rounds. Letting $x_t,\hy_t,\sc_t,\vy_t$ be the record of round $t$, we define $p_t$ to include $x_t$ but not $\hy_t,\sc_t,\vy_t$.

Then the current verifier version space is $\V_{p_t}$. 
An optimal learner needs to take the action $(\hy, \sc)$ that minimizes the maximal number of mistakes the environment will be able to enforce. The environment can enforce mistakes by combining past mistakes that happened on $p_t$ with future mistakes that may happen in a debate tree from this point on \emph{through the same function $h$}.
Thus, the learner chooses a pair $(\hy, \sc)$ in $\argmin_{\hy, \sc}Q_t(\hy, \sc)$ for

\[
Q_t(\hy, \sc) = \max_{\vy\in\{0,1\}} ~\sup_{\T\in\Delta_n(\V_{(p_t,\hy, \sc, \vy)})} ~\min_{q\in\T^\rho} ~\max_{h\in \H_{(p_t,\hy, \sc, \vy)}\cap \H_q} (\mi(h, p_t) + \mi(h,q) +\indct[h(x_t) \neq \hy]),
\]
where we use the notation $(p_t,\hy, \sc, \vy)$ for the path $p_t$ expanded with learner actions $(\hy, \sc)$ and verifier response $\vy$. 
Note that for the version spaces defined through this expanded path, the domain point component $x_{t+1}$ of the  environment response is not relevant, and we thus omit it here to not overload the notation.

\begin{lemma}\label{lemma:cooperative_deterministic_upperbound}
 Let $\H\subseteq \Y^\X$ be some hypothesis class, for which there exists a verifier class $\V$ with $\H = \H_\V$ and $\dldim(\V, n) \in \naturals$. Then, at every step $t$ of the SOA learner above, $\min_{\hy,\sc} Q_t(\hy,\sc) \leq \dldim(\V, n)$. 
  Moreover, $\dldim(\V, n)$ is also an upper bound on the total number of mistakes that the SOA learner makes during the run.
\end{lemma}

\section{Oppositional Debaters}
\label{sec:oppositional}

We now turn to consider the oppositional setting. As described in Section~\ref{sec:formal}, the standard approach from the literature (that does not involve learning) considers debaters that optimize the worst case $\tLo$ and $\fLo$ with respect to a worst-case opposing debater. However, we show here that in a learning setting, an approach minimizing a notion of \emph{regret} is more suitable. 

\subsection{Using regret to upper bound prediction mistakes}

Given $v^*$, there is a strategy for \dt that obtains $\tLo(T) = 0$, since for any $x_t$, it can choose $\ct^1,\ldots,\ct^n$ that satisfy the validation requirement of $v^*$ for any $\cf^1,\ldots,\cf^n$ (as in Definition \ref{def:verifier_validates_hypothesis}). We call a round in which \dt selected $\hy_t  = h_{v^*}(x_t)$ and such certificates an \emph{irrefutable round}. For \df, even given $v^*$, it cannot avoid incurring a loss in irrefutable rounds. Therefore, for \df a notion of regret instead of loss is relevant even in a realizable setting: Let $U(T)$ be the number of irrefutable rounds in a given interaction. The smallest possible loss for \df is therefore $U(T)$. Thus, its regret is defined as $\fRe(T) = \fLo(T) - U(T)$. For \dt in the realizable setting, its regret is equal to its loss: 
$\tRe(T) = \tLo(T)$.

The following claim shows that the total regret of the debaters upper bounds the total mistakes of the debate protocol.

\begin{claim}\label{claim:regret}
In any run of the debate protocol,  $\Mi(T) \leq \tRe(T) + \fRe(T)$.
\end{claim}
\begin{proof}
In an irrefutable round,  \dt's choices force $v^*$ to output $\vy_t = \hy_t$, regardless of the choices of \df. Since $v^*$ validates $h^* = h_{v^*}$, this means that $\hy_t=h^*(x_t)$, so there are no prediction mistakes in these rounds. Hence, $\Mi(T) \leq T-U(T)$. In every round, exactly one of the debaters incurs a loss, so
$T =  \tLo(T)+\fLo(T)$. Therefore
\[
\Mi(T) \leq \tLo(T)+\fLo(T)-U(T) = \tRe(T) + \fRe(T).
\]
As claimed.
\end{proof}

This upper bound motivates new objectives for \dt and \df in the oppositional setting when learning is involved: if each debater minimizes their worst-case \emph{regret} with respect to a worst-case opposing debater, this upper bounds the worst-case number of debate protocol mistakes: if the worst-case regret for \dt over $T$ rounds is $R_P^T$ and the worst-case regret for \df  over $T$ rounds is $R_R^T$, then for any sequence, $\Mi(T) \leq R_P^T+ R_R^T$.

Moreover, we show next that unlike the regret optimization approach, the loss optimization approach discussed above does not provide similar guarantees and can in fact completely fail. 

\subsection{Loss optimization vs.~regret optimization}

We show that even if each debater optimizes its worst-case loss against worst-case environment and opposing debater, this does not provide helpful guarantees for learning. 
First, we observe that for \df, this optimization criterion is  meaningless, since all \df strategies are optimal according to this criterion. 

\begin{claim}\label{claim:deb2}
    For any verifier class $\V$ and any number of rounds $T$, the worst-case optimal loss for $\df$ is $T$. 
\end{claim}
\begin{proof}
Consider a sequence $(x_t,\hy_t,\ct_t^1, \ldots, \ct_t^n)_{t \in [T]}$ that is consistent with some verifier $v^* \in \V$ such that for each $t$, $\forall \cf_t^1, \ldots, \cf_t^n, v^*(x_t, \ct_t^1,\cf_t^1, \ldots, \ct_t^n, \cf_t^n) = \hy_t$. Note that such $\hy_t$ and choices $\ct_t^1, \ldots, \ct_t^n$ exists for every $x \in \X$ according to the definition of a verifier. 
For this sequence, any response of \df has $\hy_t = \vy_t$. Hence, \df incurs a loss in every round, leading to $\fLo(T) = T$.
\end{proof}

In addition, in the next theorem we show that the loss of \dt cannot be used to upper bound the debate protocol mistakes $\Mi(T)$ for \emph{loss-optimizing} debaters, that is debaters that are worst-case optimal with respect to their loss as indicated above. In contrast, if the debaters are worst-case optimal with respect to their \emph{regret}, then according to the bound in Claim~\ref{claim:regret}, these regrets upper bound $\Mi(T)$. In the same example, we show that $\Mi(T)$ is large under the loss optimization scenario but small under the regret optimization scenario. 
\begin{theorem}
    There is a verifier class $\V$ such that loss-optimizing debaters lead to 
    $\ex[\Mi(T)] \geq T/4$ for all environment inputs, whereas $\ex[\tLo(T)] \leq 1.5$. On the other hand, regret-optimizing debaters lead to $\ex[\Mi(T)] \leq \ex[\tLo(T)] = 1.5$ for all environment inputs.
\end{theorem}
\begin{proof}
Let $\X = \{x\}$ and $\C = \{0,1\}$. Let $\H = \{h_0,h_1\}$, where $h_b(x) = b$ for  $b\in\{0,1\}$. Let $\V$ be the class of all verifiers $\{ v_{b,c} \mid b,c \in \{0,1\}\}$  such that $\forall \cf \in \C, v_{b,c}(x,c,\cf) = b$, and for $1-c$ we have $v_{b,c}(x,1-c,\cf) = \cf$ for $\cf \in \C$, as illustrated below.
\[
\begin{array}{c|cc}
 v_{b,c}(x,\ct,\cf) & \cf=0 & \cf=1\\
\hline
\ct=c & b & b\\
\ct = 1-c & 0 & 1
\end{array}
\]
Note that $v_{b,c}$ verifies $h_b$. 

First, consider an optimal \dt. Note that loss optimization and regret optimization are equivalent for \dt. Let $b^* \in \{0,1\},c^*\in \C$ such that $v^* = v_{b^*,c^*}$. Consider the following \dt strategy: It maintains a set of active pairs of label and certificate, initialized to $\{0,1\}\times \C$. In the first round, it selects $\hy,\ct$ uniformly at random from the set of active pairs. In each round, if it incurs a loss using some pair, it removes this pair from the active set and selects another pair for the next round uniformly at random. If it does not incur a loss in a round, it continues to use the same pair in the next round. 

To see that this is worst-case optimal, observe that whenever \dt proposes some $\hy$ and $\ct$ for some $x \in \X$ and incurs a loss with some \df certificate $\cf$, this means that 
$\vy = v_{b^*,c^*}(x,\ct,\cf) \neq \hy$. This implies that $v_{b^*, c^*} \neq v_{\hy,\ct}$, since $v_{\hy,\ct}(x,\ct,\cf) = \hy$. Hence, this \dt strategy only incurs a loss at most once for every pair until selecting $b^*, c^*$. The expected number of pair selections before selecting this pair is $1.5$, hence the expected loss of \dt is at most $1.5$. This is worst-case optimal, since any strategy that does not replace a pair that has incurred a loss will incur the loss again in the next round against a worst-case \df, thus its worst-case loss cannot be smaller.

Moreover, let $j$ be the last round before $b^*,c^*$ is found, then using this \dt strategy we get $\Mi(T) \leq j$, since no prediction mistake is made after this round. However, the value of $j$ depends on the strategy of \df, as we now show.

First, consider the optimal \df according to loss minimization. As shown in Claim~\ref{claim:deb2}, all \df strategies are optimal with respect to worst-case loss minimization. In particular, consider the strategy that in every round, upon observing $(x_t, \ct_t, \hy_t)$, selects the response $\cf_t = \hy_t$. Under this strategy, if $\ct_t = 1-c^*$, then $\vy_t = v_{b^*,c^*}(x_t,\ct_t,\cf_t) = v_{b^*,c^*}(x_t,1-c^*,\hy_t) = \hy_t$. Therefore, when the optimal \dt strategy described above is used, if in the first round \dt selects $(\hy,\ct) = (1-b^*,1-c^*)$, it incurs no loss and does not change its strategy for the rest of the run, never trying $b^*, c^*$, and incurring a prediction mistake in every round. The probability that \dt tries $(\hy,\ct) = (1-b^*,1-c^*)$ in the first round is $1/4$. Hence, using these two debaters leads to $\ex[\Mi(T)] \geq T/4$.

Second, consider the optimal \df according to regret minimization. Consider the following strategy: In every round, \df selects $\cf_t = 1-\hy_t$. Then, if $\ct_t = 1-c^*$, we have $\vy_t = v_{b^*,c^*}(x_t,\ct_t,\cf_t) = v_{b^*,c^*}(x_t,1-c^*,1-\hy_t) \neq \hy_t$ and \df does not incur a loss. If  $\ct_t = c^*$, then this is an irrefutable round, hence \df may incur a loss but it does not accumulate regret. Thus, this strategy obtains zero regret and so it is optimal with respect to regret minimization. 

Using this strategy against the optimal \dt strategy described above leads to \dt incurring a loss with every pair except for $(\hy,\ct) = (b^*,c^*)$. Therefore, the expected number of rounds until it tries $(\hy,\ct) = (b^*,c^*)$ is $1.5$, leading to $\Mi(T) \leq 1.5$. This can also be derived from Claim~\ref{claim:regret}.
\end{proof}

 The results above motivate setting the objective of oppositional debaters to regret instead of loss. Nonetheless, as we show next, even with this objective, oppositional debaters can still fail where cooperative debaters succeed.

\subsection{Gap between cooperative and oppositional debaters}

The following example shows that oppositional debaters optimizing worst-case regret can fail to obtain a low number of prediction mistakes, even when cooperative debaters succeed.

\begin{example}\label{ex:reggap}
Let $\X=\{e\}\cup\naturals$, where $e\notin\naturals$, and let
$\C=\{0,1\}$. Consider length-1 debate. For $b\in\{0,1\}$, define
\[
h_b(x)=
\begin{cases}
b,&x=e,\\
0,&x\in\naturals.
\end{cases}
\]
For each $b\in\{0,1\}$ and $s:\naturals\to\{0,1\}$, define
\[
v_{b,s}(x,p,r)=
\begin{cases}
b,&x=e,\\
0,&x\in\naturals\text{ and }(p=s(x)\text{ or }r=0),\\
1,&\text{otherwise}.
\end{cases}
\]
Set $\V=\{v_{b,s}:b\in\{0,1\},\ s:\naturals\to\{0,1\}\}$.
The following table illustrates the verifier output for $x \in \naturals$.
\[
\begin{array}{c|cc}
v_{b,s}(x,\ct,\cf) &\cf=0&\cf=1\\\hline
\ct=s(x)&0&0\\
\ct \ne s(x)&0&1
\end{array}
\]
Note that $v_{b,s}$ validate $h_b$.
\end{example}

\begin{theorem}\label{thm:reggap}
For Example~\ref{ex:reggap}, worst-case regret minimizer strategies for \df and \dt obtain $\tRe(T) \leq T/2$ and $\fRe(T) = 0$, and in the worst case, $\Mi(T) = T/2$.
On the other hand, $\dldim(\V) = 1$, hence deterministic cooperative debaters incur at most one prediction mistake. 
\end{theorem}

\section{Non-Realizability with Cooperative Debaters}

In the $k$-non-realizable setting for $k \in \naturals$, we consider the same debate protocol as described above, except that in at most $k$ rounds, the environment may provide $\vy \neq v^*(x_t, \sc_t)$. In the cooperative setting, we show upper and lower bounds for learning in the $k$-non-realizable setting. Our analysis shows that debate learning is more adversely affected by feedback noise or model misspecification than standard online learning.

\paragraph{Separation between non-realizable and realizable learning}
\label{sec:agnostic-example}
First, we show a simple verifier class that is easy to learn in the realizable cooperative case (with randomized debaters), but not when there are misspecifications.

\begin{example}\label{example:sep}
    Consider debates of length one. Let $\X=\naturals$, $\C=\{0,1\}$, and let $\ct_t$ and $\cf_t$ be the certificates at step $t$. For $b\in\{0,1\}$, let $h_b$ be the constant hypothesis $h_b(x)=b$, and let $\H=\{h_0,h_1\}$. For every bit $b$ and every function $e:\X\to \C\times\C$, define
\[
v_{b,e}(x,\ct,\cf)=
\begin{cases}
1-b, & (\ct,\cf)=e(x),\\
b, & \text{otherwise},
\end{cases}
\]
To illustrate, suppose that $e(x) = (c_0,c_1)$, then the value of $v_{b,e}(x,\ct,\cf)$ is as follows.
\[
\begin{array}{c|cc}
v_{b,e}(x,\ct,\cf) & \cf=c_1 & \cf=1-c_1\\
\hline
\ct=c_0 & 1-b & b\\
\ct=1-c_0 & b & b
\end{array}
\]
Let $\V=\{v_{b,e}:b\in\{0,1\},\ e:\X\to \C\times\C \}$. 

Clearly, for any $b,e$, $v_{b,e}$ validates $h_b$, using the $p$ that gives $b$ for any $r$. 
\end{example}

\begin{theorem}\label{thm:examplesep}
For Example~\ref{example:sep}, the following two properties hold. In the realizable case with cooperative debaters, there is a randomized strategy that obtains $\ex[\Mi(T)]<9$ for every $T$, every $v^*\in\V$ and every sequence of input instances. In the unrealizable case, for every $k\ge1$ and every $T\ge 3k$, the environment 
can choose the inputs and draw $v^*\in\V$ and the misspecified rounds at random in advance, so that the expected number of misspecified rounds is $k$ and every strategy of the debaters has $\ex[\Mi(T)]\ge\frac32k$.
\end{theorem}

\paragraph{A lower bound for general verifier classes}
Our next result shows that %
misspecification can result in a loss that is linear in $k\cdot \dldim(\V)$.
\begin{theorem}\label{thm:kdlower}
    For every $d$ and every $n$, there exists a class $\V_{d,n}$ with $\dldim(\V_d)=d$ and such that every %
    learner makes $\Omega(kd)$ mistakes in the $k$-non-realizable case.
\end{theorem}

\paragraph{An upper bound for the $k$-non-realizable setting}
We provide a matching mistake upper bound of $O(k\cdot \dldim(\V))$ for any verifier class $\V$, using a randomized strategy. \cite{oz2026discriminative} provides a general deterministic strategy for obtaining such an upper bound for interactive learning protocols. However, that strategy relies on the learner having access to its incurred loss, which is not available in debate learning. 
Our upper bound employs instead a simple randomized algorithm based on majority voting, whose structure is inspired by the Private Online Procedure of \cite{kaplan2023black}. 

\begin{theorem}\label{thm:upperkd}
    Assume the $k$-non-realizable setting. For any verifier class $\V$ with $\dldim(\V) = d \in \naturals$, there is a randomized learning strategy for cooperative debaters that incurs at most $16kd$ mistakes in expectation. 
\end{theorem}

\section{Conclusion}

In this work, we initiated the formal study of learning through debate. While debate has been used in training pipelines, no formal framework for understanding its effectiveness for \emph{learning} had been developed so far. Our analysis highlights some important considerations from a safety perspective. We show that the practice of setting the debaters up as opponents, rather than cooperating agents, can be sub-optimal. 
This finding suggests that developing training pipelines with cooperating, rather than competing agents, might be a promising avenue, which is yet underexplored.
Moreover, our results demonstrate the brittleness of learning through debate to noise in the verifier feedback or slight model misspecifications. This result shows that employing debating agents as facilitators for learning demands caution: if the verifier that these agents interact with is not fully reliable, the learning process may suffer disproportionately in accuracy.

\section*{AI Disclosure}

For this work, we used ChatGPT and Claude for brainstorming some proof ideas and some examples. All derivations and proofs in the paper were written and verified by the authors.

\section*{Acknowledgements}
We acknowledge the support of the Natural Sciences and Engineering Research Council of Canada (NSERC), [funding reference numbers RGPIN-2025-06329 and RGPIN-2024-05907]. Resources used in preparing this research were provided, in part, by the Province of Ontario, the Government of Canada through CIFAR, and companies sponsoring the Vector Institute; see \url{https://vectorinstitute.ai/partnerships/current-partners/}. This work was supported in part by a grant of access to OpenAI models through the ChatGPT for Academic Researchers program. Vinayak Pathak is supported by Advanced Research+Invention Agency (ARIA) of the United Kingdom and Coefficient Giving in San Francisco, California. Tosca Lechner was supported by a Vector Postdoctoral Fellowship.

\bibliography{references}

@article{russell2015research,
  author = {Russell, Stuart and Dewey, Daniel and Tegmark, Max},
  title = {Research Priorities for Robust and Beneficial Artificial Intelligence},
  journal = {AI Magazine},
  volume = {36},
  number = {4},
  pages = {105--114},
  year = {2015},
  doi = {10.1609/aimag.v36i4.2577}
}

@misc{amodei2016concrete,
  author = {Amodei, Dario and Olah, Chris and Steinhardt, Jacob and
            Christiano, Paul and Schulman, John and Man{\'e}, Dan},
  title = {Concrete Problems in {AI} Safety},
  year = {2016},
  howpublished = {arXiv:1606.06565},
  url = {https://arxiv.org/abs/1606.06565}
}

@techreport{openai2026huggingface,
  author = {{OpenAI}},
  title = {{OpenAI}--{Hugging Face} Incident Technical Report},
  institution = {OpenAI},
  year = {2026},
  month = aug,
}

@misc{irving2018debate,
  author = {Irving, Geoffrey and Christiano, Paul and Amodei, Dario},
  title = {{AI} Safety via Debate},
  year = {2018},
  howpublished = {arXiv:1805.00899},
  url = {https://arxiv.org/abs/1805.00899}
}

@inproceedings{browncohen2024debate,
  author = {Brown-Cohen, Jonah and Irving, Geoffrey and Piliouras, Georgios},
  title = {Scalable {AI} Safety via Doubly-Efficient Debate},
  booktitle = {Proceedings of the 41st International Conference on Machine Learning},
  series = {Proceedings of Machine Learning Research},
  volume = {235},
  pages = {4585--4602},
  year = {2024},
  publisher = {PMLR},
  url = {https://proceedings.mlr.press/v235/brown-cohen24a.html}
}

@article{vergis1986complexity,
  author = {Vergis, Anastasios and Steiglitz, Kenneth and Dickinson, Bradley},
  title = {The Complexity of Analog Computation},
  journal = {Mathematics and Computers in Simulation},
  volume = {28},
  number = {2},
  pages = {91--113},
  year = {1986},
  doi = {10.1016/0378-4754(86)90105-9}
}

@article{aaronson2005physical,
  author = {Aaronson, Scott},
  title = {{NP}-complete Problems and Physical Reality},
  journal = {ACM SIGACT News},
  volume = {36},
  number = {1},
  pages = {30--52},
  year = {2005},
  doi = {10.1145/1052796.1052804},
  url = {https://www.scottaaronson.com/papers/npcomplete.pdf}
}

@article{jumper2021alphafold,
  author = {Jumper, John and Evans, Richard and Pritzel, Alexander and
            Green, Tim and Figurnov, Michael and Ronneberger, Olaf and
            Tunyasuvunakool, Kathryn and Bates, Russ and {\v{Z}}{\'i}dek, Augustin and
            Potapenko, Anna and Bridgland, Alex and Meyer, Clemens and
            Kohl, Simon A. A. and Ballard, Andrew J. and Cowie, Andrew and
            Romera-Paredes, Bernardino and Nikolov, Stanislav and Jain, Rishub and
            Adler, Jonas and Back, Trevor and Petersen, Stig and Reiman, David and
            Clancy, Ellen and Zielinski, Michal and Steinegger, Martin and
            Pacholska, Michalina and Berghammer, Tamas and Bodenstein, Sebastian and
            Silver, David and Vinyals, Oriol and Senior, Andrew W. and
            Kavukcuoglu, Koray and Kohli, Pushmeet and Hassabis, Demis},
  title = {Highly Accurate Protein Structure Prediction with {AlphaFold}},
  journal = {Nature},
  volume = {596},
  number = {7873},
  pages = {583--589},
  year = {2021},
  doi = {10.1038/s41586-021-03819-2}
}

@incollection{yudkowsky2008artificial,
  author = {Yudkowsky, Eliezer},
  title = {Artificial Intelligence as a Positive and Negative Factor in Global Risk},
  booktitle = {Global Catastrophic Risks},
  editor = {Bostrom, Nick and {\'C}irkovi{\'c}, Milan M.},
  pages = {308--345},
  publisher = {Oxford University Press},
  year = {2008},
  doi = {10.1093/oso/9780198570509.003.0021}
}

@inproceedings{soares2015corrigibility,
  author = {Soares, Nate and Fallenstein, Benja and Yudkowsky, Eliezer and
            Armstrong, Stuart},
  title = {Corrigibility},
  booktitle = {Workshops at the Twenty-Ninth {AAAI} Conference on Artificial Intelligence},
  year = {2015},
  url = {https://intelligence.org/wp-content/uploads/2024/10/Corrigibility.pdf}
}

@inproceedings{everitt2017corrupted,
  author = {Everitt, Tom and Krakovna, Victoria and Orseau, Laurent and Legg, Shane},
  title = {Reinforcement Learning with a Corrupted Reward Channel},
  booktitle = {Proceedings of the Twenty-Sixth International Joint Conference on Artificial Intelligence},
  pages = {4705--4713},
  year = {2017},
  doi = {10.24963/ijcai.2017/656}
}

@incollection{soares2017agent,
  author = {Soares, Nate and Fallenstein, Benya},
  title = {Agent Foundations for Aligning Machine Intelligence with Human Interests: A Technical Research Agenda},
  booktitle = {The Technological Singularity: Managing the Journey},
  series = {The Frontiers Collection},
  pages = {103--125},
  publisher = {Springer},
  year = {2017},
  doi = {10.1007/978-3-662-54033-6_5}
}

@inproceedings{armstrong2018occam,
  author = {Armstrong, Stuart and Mindermann, S{\"o}ren},
  title = {{Occam's} Razor Is Insufficient to Infer the Preferences of Irrational Agents},
  booktitle = {Advances in Neural Information Processing Systems},
  volume = {31},
  year = {2018},
  publisher = {Curran Associates, Inc.},
  url = {https://proceedings.neurips.cc/paper/2018/hash/d89a66c7c80a29b1bdbab0f2a1a94af8-Abstract.html}
}

@incollection{soares2018value,
  author = {Soares, Nate},
  title = {The Value Learning Problem},
  booktitle = {Artificial Intelligence Safety and Security},
  editor = {Yampolskiy, Roman V.},
  pages = {89--97},
  publisher = {Chapman and Hall/CRC},
  year = {2018},
  doi = {10.1201/9781351251389-7}
}

@misc{hubinger2019risks,
  author = {Hubinger, Evan and van Merwijk, Chris and Mikulik, Vladimir and
            Skalse, Joar and Garrabrant, Scott},
  title = {Risks from Learned Optimization in Advanced Machine Learning Systems},
  year = {2019},
  howpublished = {arXiv:1906.01820},
  url = {https://arxiv.org/abs/1906.01820}
}

@incollection{taylor2020alignment,
  author = {Taylor, Jessica and Yudkowsky, Eliezer and LaVictoire, Patrick and
            Critch, Andrew},
  title = {Alignment for Advanced Machine Learning Systems},
  booktitle = {Ethics of Artificial Intelligence},
  editor = {Liao, S. Matthew},
  pages = {342--382},
  publisher = {Oxford University Press},
  year = {2020},
  doi = {10.1093/oso/9780190905033.003.0013}
}

@inproceedings{skalse2022reward,
  author = {Skalse, Joar and Howe, Nikolaus and Krasheninnikov, Dmitrii and
            Krueger, David},
  title = {Defining and Characterizing Reward Gaming},
  booktitle = {Advances in Neural Information Processing Systems},
  volume = {35},
  pages = {9460--9471},
  year = {2022},
  publisher = {Curran Associates, Inc.},
  doi = {10.52202/068431-0687}
}

@article{irving2019social,
  author = {Irving, Geoffrey and Askell, Amanda},
  title = {{AI} Safety Needs Social Scientists},
  journal = {Distill},
  volume = {4},
  number = {2},
  year = {2019},
  doi = {10.23915/distill.00014}
}

@inproceedings{parrish2022single,
  author = {Parrish, Alicia and Trivedi, Harsh and Perez, Ethan and Chen, Angelica and
            Nangia, Nikita and Phang, Jason and Bowman, Samuel R.},
  title = {Single-Turn Debate Does Not Help Humans Answer Hard Reading-Comprehension Questions},
  booktitle = {Proceedings of the First Workshop on Learning with Natural Language Supervision},
  pages = {17--28},
  year = {2022},
  publisher = {Association for Computational Linguistics},
  doi = {10.18653/v1/2022.lnls-1.3}
}

@inproceedings{parrish2022two,
title={Two-Turn Debate Does Not Help Humans Answer Hard Reading Comprehension Questions},
author={Alicia Parrish and Harsh Trivedi and Nikita Nangia and Jason Phang and Vishakh Padmakumar and Amanpreet Singh Saimbhi and Samuel R. Bowman},
booktitle={NeurIPS ML Safety Workshop},
year={2022},
url={https://openreview.net/forum?id=9wLAwDrYDsQ}
}

@misc{browncohen2025prover,
  author = {Brown-Cohen, Jonah and Irving, Geoffrey and Piliouras, Georgios and
            Chen, Lijie and Li, Jiawei and Xun, Zhiyang},
  title = {Avoiding Obfuscation with Prover-Estimator Debate},
  year = {2025},
  howpublished = {arXiv:2506.13609},
  url = {https://arxiv.org/abs/2506.13609}
}

@inproceedings{
oz2026discriminative,
title={Discriminative Feature Feedback with General Teacher Classes},
author={Bar Oz, Omri and Lechner, Tosca and Sabato, Sivan},
booktitle={37th International Conference on Algorithmic Learning Theory},
year={2026},
url={https://openreview.net/forum?id=jCWFarqEpZ}
}

@article{kaplan2023black,
  title={Black-box differential privacy for interactive ml},
  author={Kaplan, Haim and Mansour, Yishay and Moran, Shay and Nissim, Kobbi and Stemmer, Uri},
  journal={Advances in Neural Information Processing Systems},
  volume={36},
  pages={77313--77330},
  year={2023}
}

@inproceedings{FilmusHaMeMo23,
  title={Optimal prediction using expert advice and randomized littlestone dimension},
  author={Filmus, Yuval and Hanneke, Steve and Mehalel, Idan and Moran, Shay},
  booktitle={The Thirty Sixth Annual Conference on Learning Theory},
  pages={773--836},
  year={2023},
  organization={PMLR}
}

@inproceedings{BenDavidPaSh09,
  title={Agnostic Online Learning.},
  author={Ben-David, Shai and P{\'a}l, D{\'a}vid and Shalev-Shwartz, Shai},
  booktitle={COLT},
  volume={3},
  pages={1},
  year={2009}
}

@inproceedings{AlonBeDaMoNaYo21,
  title={Adversarial laws of large numbers and optimal regret in online classification},
  author={Alon, Noga and Ben-Eliezer, Omri and Dagan, Yuval and Moran, Shay and Naor, Moni and Yogev, Eylon},
  booktitle={Proceedings of the 53rd annual ACM SIGACT symposium on theory of computing},
  pages={447--455},
  year={2021}
}

@inproceedings{
chen2023playing,
title={Playing Large Games with Oracles and {AI} Debate},
author={Xinyi Chen and Angelica Chen and Dean Foster and Elad Hazan},
booktitle={Agentic Markets Workshop at ICML 2024},
year={2024},
url={https://openreview.net/forum?id=b2gfTWXG7V}
}

@misc{buzaglo2025hidden,
  author = {Buzaglo, Gon and Golowich, Noah and Hazan, Elad},
  title = {The Hidden Game Problem},
  year = {2025},
  howpublished = {arXiv:2510.03845},
  url = {https://arxiv.org/abs/2510.03845}
}

@article{freund1999adaptive,
  author = {Freund, Yoav and Schapire, Robert E.},
  title = {Adaptive Game Playing Using Multiplicative Weights},
  journal = {Games and Economic Behavior},
  volume = {29},
  number = {1--2},
  pages = {79--103},
  year = {1999},
  doi = {10.1006/game.1999.0738}
}

@article{hart2000simple,
  author = {Hart, Sergiu and Mas-Colell, Andreu},
  title = {A Simple Adaptive Procedure Leading to Correlated Equilibrium},
  journal = {Econometrica},
  volume = {68},
  number = {5},
  pages = {1127--1150},
  year = {2000},
  doi = {10.1111/1468-0262.00153}
}

@misc{arnesen2024training,
  author = {Arnesen, Samuel and Rein, David and Michael, Julian},
  title = {Training Language Models to Win Debates with Self-Play
           Improves Judge Accuracy},
  year = {2024},
  howpublished = {arXiv:2409.16636},
  url = {https://arxiv.org/abs/2409.16636}
}

\clearpage
\appendix

\section{Deferred proofs}

\begin{proof}[Proof of Lemma~\ref{lemma:cooperative_deterministic_upperbound}]
For an instance $x$, define the set of trees $\Delta_n(\V,x)$ to be the set of trees in $\Delta_n(\V)$ that have $x$ at the root.
Define the set of trees $\Delta_n(\V,x,\hy,\sc)$ to be the set of trees that result from taking any tree from $\Delta_n(\V,x)$ and removing all the edges (and corresponding sub-trees) coming out of $x$ except the one labeled $\hy,\sc$. 

First, we show that $\min_{\hy,\sc}Q_t(\hy,\sc)$ is non increasing in $t$. Consider a round $t$ in some run of SOA. We have
\begin{align*}
  &\min_{\hy,\sc}Q_t(\hy,\sc) = \min_{\hy,\sc} \max_{\vy\in\{0,1\}} ~\sup_{\T\in\Delta_n(\V_{(p_t,\hy, \sc, \vy)})} ~\min_{q\in\T^\rho} ~\max_{h\in \H_{(p_t,\hy, \sc, \vy)}\cap \H_q} (\mi(h, p_t) + \mi(h,q)+\indct[h(x_t) = \hy])\\
  &\leq
    \min_{\hy,\sc}  ~\sup_{\T\in\Delta_n(\V_{p_t},x_t,\hy,\sc)} ~\min_{q\in\T^\rho} ~\max_{h\in \H_{p_t}\cap \H_q} (\mi(h, p_t) + \mi(h,q))
\end{align*}
For a path $q\in\T^\rho$, denote $f(q) := \max_{h\in \H_{p_t}\cap \H_q} (\mi(h, p_t) + \mi(h,q))$, and let
\[
q_\T \in \argmin_{q\in\T^\rho} f(q).
\]
denote the path in tree $\T$ that minimizes the sum of mistakes an $h$ can incur in $q_\T$ and $p_t$ combined.
Note that if $\T \in \Delta_n(\V_{p_t},x_t,\hy,\sc)$ then $q_\T$ must start with $(x_t,\hy,\sc)$. Let $q'_\T$ be the rest of the path of $q_\T$ and write $q_\T = (x_t,\hy,\sc);q'_\T$. 
Then we have 
\begin{align*}
&\min_{\hy,\sc}Q_t(\hy,\sc)  \leq \min_{\hy,\sc}  ~\sup_{\T\in\Delta_n(\V_{p_t},x_t,\hy,\sc)} f(q_\T) = 
\min_{\hy,\sc}\sup_{\T\in\Delta_n(\V_{p_t},x_t,\hy,\sc)} f((x_t,\hy,\sc);q'_\T) =: R
\end{align*}

Define $\T_{\hy,\sc} \in \argmax_{\T \in \Delta_n(\V_{p_t},x_t,\hy,\sc)}f((x_t,\hy,\sc);q'_\T)$ (if a maximum exists)\footnote{A similar argument can be made if there is no maximum, by using a sequence of $\T_{\hy,\sc}$ choices that are larger than a sequence converging to the supremum}
and let $\T^*$ be the tree constructed by joining each of the trees $\T_{\hy,\sc}$ at the root $x_t$. Note that $\T^* \in \Delta_n(\V_{p_t},x_t)$. Let $X_{\hy,\sc}$ be the set of paths that start with $(x_t, \hy,\sc)$.
Then
\[
\min_{q \in (\T^*)^\rho} f(q) = \min_{\hy,\sc} \min_{q \in \T^*\cap X_{\hy,\sc}} f(q) = \min_{\hy,\sc}f((x_t,\hy,\sc);q'_{\T_{\hy,\sc}}) = R.
\]
Hence,
\[
\sup_{\T \in \Delta_n(\V_{p_t},x_t)} \min_{q \in \T^\rho} f(q)  \geq \min_{q \in (\T^*)^\rho} f(q) \geq R \geq \min_{\hy,\sc}Q_t(\hy,\sc).
\]
We now upper bound the LHS.
\begin{align*}
&\sup_{\T \in \Delta_n(\V_{p_t},x_t)} \min_{q \in \T^\rho} f(q) = \sup_{\T\in\Delta_n(\V_{p_t},x_t)} ~\min_{q\in\T^\rho} ~\max_{h\in \H_{p_t}\cap \H_q} (\mi(h, p_{t}) + \mi(h,q))
  \\
  &\quad= \sup_{\T\in\Delta_n(\V_{(p_{t-1},\hy_{t-1}, \sc_{t-1}, \vy_{t-1})},x_t)} ~\min_{q\in\T^\rho} ~\max_{h\in \H_{(p_{t-1},\hy_{t-1}, \sc_{t-1}, \vy_{t-1})}\cap \H_q}
  \\
  &\qquad\qquad (\mi(h, p_{t-1}) + \mi(h,q)+\indct[h(x_{t-1}) = \hy_{t-1}])
  \\
  &\quad\leq \max_{\vy\in\{0,1\}} ~\sup_{\T\in\Delta_n(\V_{(p_{t-1},\hy_{t-1}, \sc_{t-1}, \vy)})} ~\min_{q\in\T^\rho} ~\max_{h\in \H_{(p_{t-1},\hy_{t-1}, \sc_{t-1}, \vy)}\cap \H_q}
  \\
  &\qquad\qquad (\mi(h, p_{t-1}) + \mi(h,q)+\indct[h(x_{t-1}) = \hy_{t-1}])
\\   
  &\quad=\min_{\hy,\sc}\max_{\vy\in\{0,1\}} ~\sup_{\T\in\Delta_n(\V_{(p_{t-1},\hy, \sc, \vy)})} ~\min_{q\in\T^\rho} ~\max_{h\in \H_{(p_{t-1},\hy, \sc, \vy)}\cap \H_q} (\mi(h, p_{t-1}) + \mi(h,q)+\indct[h(x_{t-1}) = \hy]) \\
  &\quad= \min_{\hy,\sc}Q_{t-1}(\hy,\sc).
\end{align*}
We conclude that $\min_{\hy,\sc}Q_{t}(\hy,\sc) \leq \min_{\hy,\sc}Q_{t-1}(\hy,\sc)$. Now consider $Q_1$. By definition, $\mi(h, p_1) = 0$ since $p_1$ includes no hypothesized labels, and 
similarly to the above,
\[
\min_{\hy,\sc}Q_1(\hy, \sc) = \max_{\vy\in\{0,1\}} ~\sup_{\T\in\Delta_n(\V_{(p_1,\hy, \sc, \vy)})} ~\min_{q\in\T^\rho} ~\max_{h\in \H_{(p_1,\hy, \sc, \vy)}\cap \H_q} (\mi(h,q) +\indct[h(x_t) \neq \hy]).
\]
Since $\V_{p_1} = \V$, we get
\begin{align*}
\min_{\hy,\sc}Q_1(\hy, \sc) &\leq \sup_{\T\in\Delta_n(\V_{p_1},x_1)} ~\min_{q\in\T^\rho} ~\max_{h\in \H_{p_1}\cap \H_q} (\mi(h, p_{1}) + \mi(h,q)) \\
&= \sup_{\T\in\Delta_n(\V,x_1)} ~\min_{q\in\T^\rho} ~\max_{h\in \H_q} (\mi(h,q)) \leq \dldim(\V, n).
\end{align*}
The latter follows from the definition of $\dldim(\V, n)$.
Lastly, to upper bound the total number of mistakes of the SOA run, note that for path $p_t$ taken by SOA, at any round $t$,
\begin{align*}
&\max_{h\in \H_{p_t}} (\mi(h, p_t)) \\
&\quad\leq
\min_{\hy,\sc} \max_{\vy\in\{0,1\}} ~\sup_{\T\in\Delta_n(\V_{(p_{t-1},\hy, \sc, \vy)})} ~\min_{q\in\T^\rho} ~\max_{h\in \H_{(p_{t-1},\hy, \sc, \vy)}\cap \H_q} (\mi(h, p_{t-1}) + \mi(h,q)+\indct[h(x_t) = \hy]) \\
&\quad= \min_{\hy,\sc}Q_{t-1}(\hy, \sc) \leq \dldim(\V, n).
\end{align*}

The first inequality follows by the realizability assumption, which implies that there is at least one $q$ with a non-empty $\H_q$. 
\end{proof}

\begin{proof}[Proof of Theorem~\ref{thm:reggap}]
Consider a cooperative learner: it can predict $\hy = 0$ for any $x_t \in \naturals$ without making any mistake. For $x_t = e$, it needs to make at most one mistake, since for $x_t = e$, $\vy = b$ is provided, revealing the label of $e$.

On the other hand, consider possible worst-case regret-optimizing debaters. 
First, define a regret-optimizing strategy for \df. This strategy refutes whenever possible: for any input $(x_t,\hy_t,\ct_t)$, it selects $\ct_t = 1-\hy_t$. Whenever $x \in \naturals$ and $\ct_t \neq s(x)$, this gives $\vy_t = v_{b,s}(x_t,\ct_t,\cf_t) = \cf_t \neq \hy_t$. Thus, its regret is zero and so optimal. 

Second, define a regret-optimizing strategy for \dt. For $x_t = e$, this strategy predicts a uniformly random label with $\ct_t = 0$. For $x_t \in \naturals$, it predicts $\hy = 0$ with a uniformly random $\ct_t$. Its expected loss on a given sequence is equal to half the number of rounds in which $e$ appears, thus $\tLo(T) = \tRe(T) \leq T/2$ and $T/2$ is also its worst-case regret over all sequences. This strategy is worst-case regret optimal, because any strategy obtains a regret of $T/2$ in the following scenario: If the environment selects $v^*$ uniformly at random, and presents distinct $x_t \in \naturals$ in each round, and if \df always selects $\ct_t = 1$, then \df always has a probability of $1/2$ to select $\hy_t = \indct[\ct_t\neq s(x_t)] = v^*(x_t, \ct_t, 1) = \vy_t$. Thus it suffers a loss and regret of $T/2$ in expectation. 

Now, observe that running these regret-optimizing \dt and \df against each other with input $e$ results in a prediction mistake. Hence, if $x_t = e$ in every round, $\Mi(T) = T/2$. 
\end{proof}

\begin{proof}[Proof of Theorem~\ref{thm:examplesep}]
We will consider debates of length one. Let $\X=\naturals$, $\C=\{0,1\}$, and let $\ct_t$ and $\cf_t$ be the certificates at step $t$. For $b\in\{0,1\}$, let $h_b$ be the constant hypothesis $h_b(x)=b$, and let $\H=\{h_0,h_1\}$. For every bit $b$ and every function $e:\X\to \C\times\C$, define
\[
v_{b,e}(x,\ct,\cf)=
\begin{cases}
1-b, & (\ct,\cf)=e(x),\\
b, & \text{otherwise},
\end{cases}
\]
To illustrate, suppose that $e(x) = (c_0,c_1)$, then the value of $v_{b,e}(x,\ct,\cf)$ is as follows.
\[
\begin{array}{c|cc}
v_{b,e}(x,\ct,\cf) & \cf=c_1 & \cf=1-c_1\\
\hline
\ct=c_0 & 1-b & b\\
\ct=1-c_0 & b & b
\end{array}
\]
Let $\V=\{v_{b,e}:b\in\{0,1\},\ e:\X\to \C\times\C$\}. 

Clearly, for any $b,e$, $v_{b,e}$ validates $h_b$, using the $p$ that gives $b$ for any $r$. %
Next, we give a randomized strategy for learning $h^*$ in the realizable case. Fix $v^*\in\V$ and a sequence of inputs, and let $h^*=h_{v^*}=h_b$.

In each round, both \dt and \df choose their arguments $\ct,\cf$ uniformly and independently. \dt proposes as a label the majority of the earlier feedback $\vy_1,\ldots,\vy_{t-1}$, breaking ties arbitrarily. 
The values $\vy_t$ are independent, and each value equals $b$ with probability $3/4$. After $s$ rounds, the majority is wrong only if at most $s/2$ of the $s$ values equal $b$, while their expected number is $3s/4$. By Hoeffding's inequality, this has probability at most $e^{-2(s/4)^2/s}=e^{-s/8}$. Summing over rounds gives $\ex[\Mi(T)]\le\sum_{s\ge0}e^{-s/8}=1/(1-e^{-1/8})<9$.

Next, we prove a lower bound in the unrealizable case. Suppose that $v^*$ is selected such that for each instance, $1-b$ appears at a uniformly random entry of the response matrix. Consider a sequence of non-repeating instances $x_1,x_2,\ldots x_T = 1,2,\ldots, T$, and assume that each round is misspecified with probability $1/3$, and that in such a round the response is always $1-b$ for any $\ct,\cf$. In rounds that are not misspecified, the probability of observing $b$ is $3/4$, while it is zero in misspecified rounds. Hence, the probability of observing $b$ is exactly $1/2$. It is thus impossible to gain any information on $b$ from the interaction, which implies that any prediction strategy of $\ct$ will result in an expected mistake rate of $1/2$. Since the rate of misspecifications is $1/3$, the expected number of misspecification rounds is $k = T/3$. Thus, the expected mistake rate is $\frac{3}{2}k$. 
\end{proof}

\begin{proof}[Proof of Theorem~\ref{thm:kdlower}]

Let $\mathcal{X}=\{x_0\}$ and $\C = [4d]$.
Define the following verifiers. Let $v_0: \X \times \C^2 \to \{0,1\}$ defined by  $v_0(x_0,c_1,c_2) = 0 $ for every $c_1, c_2 \in \C$.
Furthermore for every $j \in [4d]$, let $v_j: \X \times \C^2 \to \{0,1\}$ be defined by 
\[v_j (x_0, c_1, c_2) = \begin{cases}
    1 & \text{ if } c_1=j \\
    1 & \text{ if } c_2=j \\
        0 & \text{ otherwise. } \\
\end{cases}\]

Now let $\mathcal{V}_d= \{v_j : j \in [4d] \} \cup \{v_0\}$.

We note that $v_0$ is a verifier for the all-zero hypothesis $h_0: \X \to \{0,1\}$, $h_0(x_0) = 0$.
Furthermore, we note that every $v_j$ is a verifier for the all-one hypothesis $h_1: \X \to \{0,1\}$, $h_1(x_0) = 1$. We set $\H = \{h_0, h_1\}$.

\begin{claim}
 $\H$ is learnable through debate of length $1$ through access to $\mathcal{V}_d$ in the realizable case with a mistake bound of $d$.
\end{claim}
We will argue that the following learning strategy is successful: In round $t$, predict the label $\hat{y}_t= 1$ if $t$ is odd and $\hat{y}_t= 0$ if $t$ is even. As certificates $(c_1^t, c_2^t)$ choose $(2t-1,2t)$.
After $2d$ rounds, if the learner has seen one round $r$ with verifier feedback $\tilde{y}_r = 1$, keep predicting $\hat{y}_t = 1$ for $t > 2d$.
After $2d$ rounds, if the learner has only seen verifier feedback $\tilde{y}_1 = \dots = \tilde{y}_{2d} = 0$, keep predicting $\tilde{y}_t = 0$ for $t > 2d$.

We are in the realizable case, so we know the true verifier is either $v_j$ for some $j\in [4d]$ or it is $v_0$.

If the true verifier is $v^* = v_j$ for some $j\in [4d]$, then in one of the rounds from $1$ to $2d$, i.e., round $\lceil \frac{j}{2}\rceil$, the verifier feedback will have been $\tilde{y}_{\lceil\frac{j}{2}\rceil} =1$.
In this case the learner will have made $d$ mistakes in rounds $1$ to $2d$ and no subsequent mistakes as the learner kept predicting $1$ after $2d$ rounds and the true verifier $v_j$ verifies the hypothesis $h_1$.

If the true verifier is $v^* = v_0$, then the true hypothesis is $h_0$. Furthermore, in rounds from $1$ to $2d$ the verifier feedback received by the learner is $\vy_1 = \dots = \vy_{2d} = 0$.
In this case the learner will have made $d$ mistakes in rounds $1$ to $2d$ and no subsequent mistakes as the learner kept predicting $0$ after $2d$ which is consistent with the true hypothesis $h_0$.

Thus this learning strategy makes at most $d$ mistakes.

\begin{claim}
    Every randomized learner for $\H$ through access to $\V_d$ via debate of length $1$ makes at least $\Omega(kd)$ mistake in expectation. 
\end{claim}

To show this we will fix a randomized adversary that picks the true verifier in advance, and can decide to give a different response in up to $k$ rounds after having seen the learners action $(\hat{y}_t, c_1^t, c_2^t)$ for that round. 
The adversary picks $v^*$ as $v_0$ with probability $1/2$ and for every $j\in [4d]$ picks every $v_j$ with probability $1/8d$.
Thus the true hypothesis is $h^* = h_0$ with probability $1/2$ and $h^* = h_1$ with probability $1/2$.

Now for any sequence, the adversary will flip to $0$ the first $k$ verifier responses that would have been $1$ if it responded according to the true verifier.

Now consider any fixed (potentially randomized) learning  strategy $\mathcal{A}: (\X \times \Y\times \C^2 \times \Y)^*\times \X \to \Y \times \C^2$, that in every round $t$ maps the history of previous examples
\[
H = ((x_1, \hy_1, \sc_1, \vy_1 ), \dots, (x_{t-1}, \hy_{t-1}, \sc_{t-1}, \vy_{t-1} ))
\]
and current example $x_t$ to the next action $(\hy_t, \bar{c}_t)$.

We will call a history $H_{t}$ with $\vy_1 = \dots  = \vy_{d-1} = 0$ a \emph{null-history}. We will analyse both the learners behaviour given a null-history and the likelihood of obtaining a null-history in round $t$. We will combine these observations in order to show that the likelihood of obtaining a null-history is high and that when given a null-history in most cases the learner will not be able to successfully distinguish between verifiers for hypothesis $h_0$ and verifiers for hypothesis $h_1$.

For every $j$, we consider the count $N_j(t) = |\{\bar{c}_r: 1\leq r \leq t-1 \text{ with } c_1^t=j \text{ or } c_2^t=j\}|$.

We note that given a null-history, if $v^* = v_0$, then independent of the learner's output in round $t$, we have $\ty_t = 0$ with probability $1$.
Furthermore, given a null history if $v^* = v_j$ then the adversary will output
\[ \vy_t = \begin{cases}
    0 &\text{if } N_{j}(t) \leq k \\
        1 &\text{otherwise.}
\end{cases}\]
with probability $1$.

Thus, the only case in which $H_t$ is a nullhistory but $H_{t+1}$ is not a null-history is the case where
$v^* = v_j$ and $ N_j(t) = k $.
It follows that $H_t$ is a nullhistory unless $v^* =v_j$ with $N_j(t) >k$.

We note that for every sequence of $t$ rounds, we have
\[
\sum_{j=1}^{4d} N_j(t) \leq 2t-2,
\]
since in every round $r \in [t]$ the learner can only increase the count for two different $j_1,j_2$ , namely $j_1= c_1^r$ and $j_2 = c_2^r$. 
Thus,
\[
\sum_{j=1}^{4d} \indct[N_j(t) > k] \leq \lfloor \frac{2t}{k} \rfloor
\]

Thus, after $t$ rounds for every sequence of choices by the learner, there are at most $\frac{2t}{k}$ different possible verifiers from $\{v_j: j\in [4d]\}$ that if the adversary picked it as the true verifier, it resulted in interactions between adversary and learner results in a history $H_t$ that is not a null-history.

The probability that the adversary picked such $v_j$ is therefore at most
\[ \frac{\frac{2t}{k}}{8d} = \frac{t}{4kd}.\]

We note, that given a null-history $H_t$ every learner makes a mistake with probability $1/2$ in the next round. That is because if the precition is $0$, it will fail in the case in which $v^*= v_j$ for $j\in [4d]$, which happens with probability $1/2$. And if the learner predicts $1$, it will fail in the case in which $v^* =v_0$, which happens with probability $1/2$.
Thus for every round $t$, 
\[\mathbb{E} [\indct[\hy_r \neq h_v(x)]|H_r \text{ is nullhistory}]  \geq 1/2\]

Using the probability of the sequence being a non-nullhistory at time $t$ together with the above inequality for the prediction error, given that the history is a nullhistory, we obtain the following:

\begin{align*}
   &\mathbb{E}[\sum_{r=1}^t  \indct[\hy_r \neq h_v(x)]]\\
   &\geq
   \mathbb{E} [\sum_{r=1}^t  \indct[\hy_r \neq h_v(x)|H_r \text{ is nullhistory}] \indct[H_r \text{ is nullhistory}]] \\
 & \geq \sum_{r=1}^t \frac{1}{2} \cdot \left(1-\frac{r}{4kd} \right)
   \geq \frac{t}{2}-\frac{t(t+1)}{16kd}  = \frac{t(8dk -t-1)}{16kd}.
\end{align*}

For $t = 4kd $ the above evaluates to $kd -\frac{1}{4} \in \Omega(kd)$. This proves the lower bound.
\end{proof}

\begin{proof}[Proof of Theorem~\ref{thm:upperkd}]
The strategy works as follows: 
\begin{itemize}
    \item Initialize $4k$ copies of SOA for $\V$.
    \item In each round $t$
    \begin{itemize}
        \item Set $\hy_t$ to the majority label proposed by the $4k$ copies.
        \item Set $\sc_t$ to the certificate $\sc$ suggested by one of the copies $i_t$, where $i_t$ selected uniformly at random from $[4k]$. Get $\vy_t$ and feed it to the same copy.
    \end{itemize}
\end{itemize}
Call a copy ``uncorrupted at time $t$'' if it has not been provided with a corrupted feedback until round $t$. Since there can be at most $k$ misspecified rounds, at least $3k$ copies are uncorrupted at each time $t$. For round $t$, let $M_t = \indct[\hy_t \neq y_t]$. Let $\hy_t^i$ be the label proposed by copy $i$ at round $t$, and define $Z_t = \indct[\hy_t^{i_t} \neq y_t \wedge i_t \text{ is uncorrupted at time }$t$]$. Any copy $i_t$ can make at most $d$ mistakes in rounds in which its copy is updated, as long as it is uncorrupted. Hence, $\sum_t Z_t \leq 4kd$. On the other hand, if $M_t = 1$, then at least $2k$ copies $i$ have $\hy_t^{i} \neq y_t$ in round $t$, and at least $k$ of these are uncorrupted at time $t$. 
Therefore, for any history $H_t$ of the run until round $t$, $\Pr(Z_t = 1 \mid M_t = 1, H_t) \geq 1/4$. 
Thus,
\[
\ex[Z_t \mid H_t] \geq \Pr(Z_t = 1 \mid M_t = 1, H_t)\cdot \Pr(M_t = 1 \mid H_t) \geq \Pr(M_t = 1 \mid H_t)/4 = \ex[M_t \mid H_t]/4.
\]
Taking expectation over $H_t$, we get $\ex[Z_t] \geq \ex[M_t = 1]/4$.
Therefore,
\[
4kd \geq \ex[\sum_t Z_t] = \ex[\sum_t M_t]/4.
\]
It follows that $\ex[\sum_t M_t] \leq 16kd$.
\end{proof}

\end{document}